\documentclass[a4paper,conference]{IEEEtran}
\IEEEoverridecommandlockouts
\usepackage{cite}
\usepackage{amsmath,amssymb,amsfonts,empheq}
\usepackage{algorithmic}
\usepackage{graphicx}
\usepackage{rotating}
\usepackage{textcomp}
\usepackage{sansmath}
\usepackage{pgfplots}
\usepackage{multirow}
\usepackage{tikz}
\usepackage{mathtools}
\usepackage{dsfont}
\usepackage{xcolor}
\pgfplotsset{compat=1.14}
\def\BibTeX{{\rm B\kern-.05em{\sc i\kern-.025em b}\kern-.08em
    T\kern-.1667em\lower.7ex\hbox{E}\kern-.125emX}}
\usepackage{hyperref}
\hypersetup{
    colorlinks=true,
    linkcolor=red,
    filecolor=magenta,      
    urlcolor=cyan,
}

\begin{document}
\setlength{\abovedisplayskip}{3pt}
\setlength{\belowdisplayskip}{5pt}
\title{Polyp Detection and Segmentation using Mask R-CNN: Does a Deeper Feature Extractor CNN Always Perform Better?
\thanks{This work was supported by Research Council of Norway through the industrial Ph.D. project under the contract number 271542/O30.}
}

\author{
    \IEEEauthorblockN{Hemin Ali Qadir\textsuperscript{1,2,5}, Younghak Shin\textsuperscript{6}, Johannes Solhusvik\textsuperscript{2,5}, Jacob Bergsland\textsuperscript{1},} 
    \IEEEauthorblockN{Lars Aabakken\textsuperscript{1,4}, Ilangko Balasingham\textsuperscript{1,3}}
    \vspace{0.3cm}
    \IEEEauthorblockA{\textit{\textsuperscript{1}Intervention Centre, Oslo University Hospital, Oslo, Norway}}
    \IEEEauthorblockA{\textit{\textsuperscript{2}Department of Informatics, University of Oslo, Oslo, Norway}}
    \IEEEauthorblockA{\textit{\textsuperscript{3}Department of Electronic Systems, Norwegian University of Science and Technology, Trondheim, Norway}}
    
    \IEEEauthorblockA{\textit{\textsuperscript{4}Department of Transplantation Medicine, University of Oslo, Oslo, Norway}}
    \IEEEauthorblockA{\textit{\textsuperscript{5}OmniVision Technologies Norway AS, Oslo, Norway}}
    \IEEEauthorblockA{\textit{\textsuperscript{6}LG CNS, Seoul, Korea}}
}

\maketitle

\begin{abstract}
Automatic polyp detection and segmentation are highly desirable for colon screening due to polyp miss rate by physicians during colonoscopy, which is about 25\%. However, this computerization is still an unsolved problem due to various polyp-like structures in the colon and high interclass polyp variations in terms of size, color, shape and texture. In this paper, we adapt Mask R-CNN and evaluate its performance with different modern convolutional neural networks (CNN) as its feature extractor for polyp detection and segmentation. We investigate the performance improvement of each feature extractor by adding extra polyp images to the training dataset to answer whether we need deeper and more complex CNNs, or better dataset for training in automatic polyp detection and segmentation. Finally, we propose an ensemble method for further performance improvement. We evaluate the performance on the 2015 MICCAI polyp detection dataset. The best results achieved are 72.59\% recall, 80\% precision, 70.42\% dice, and  61.24\% jaccard. The model achieved state-of-the-art segmentation performance.  
\end{abstract}


\section{Introduction}\label{sec:Intro}
Colorectal cancer is the second most common cause of cancer-related death in the United States for both men and women, and its incidence increases every year \cite{b1}. Colonic polyps, growths of glandular tissue at colonic mucosa, are the major cause of colorectal cancer. Although they are initially benign, they might become malignant over time if left untreated \cite{b2}. Colonoscopy is the primary method for screening and preventing polyps from becoming cancerous \cite{b3}. However, colonoscopy is dependent on highly skilled endoscopists and high level of eye-hand coordination, and recent clinical studies have shown that 22\%--28\% of polyps are missed in patients undergoing colonoscopy \cite{b4}.

Over the past decades, various computer aided diagnosis systems have been developed to reduce polyp miss rate and improve the detection capability during colonoscopy \cite{b5,b6,b7,b8,b9,b10,b11,b12,b13,b14,b15,b16,b17,b18, b19}. The existing automatic polyp detection and segmentation methods can be roughly grouped into two categories: 1) those which use hand-crafted features \cite{b5,b6,b7,b8,b9,b10,b11}, 2) those which use data driven approach, more specifically deep learning method \cite{b12,b13,b14,b15,b16,b17,b18}. 

The majority of hand-crafted based methods can be categorized into two groups: texture/color based \cite{b5,b6,b7,b8} and shape based \cite{b9,b10,b11}. In \cite{b5,b6,b7,b8}, color wavelet, texture, Haar, histogram of oriented gradients and local binary pattern were investigated to differentiate polyps from the normal mucosa. Hwang et al. \cite{b9} assumed that polyps have elliptical shape that distinguishes polyps from non-polyp regions. Bernal et al. \cite{b10} used valley information based on polyp appearance to segment potential regions by watersheds followed by region merging and classification. Tajbakhsh et al. \cite{b11} used edge shape and context information to accumulate votes for polyp regions. These feature patterns are frequently similar in polyp and polyp-like normal structures, resulting in decreased performance. 

To overcome the shortcomings of the hand-crafted features, a data driven approach based on CNN was proposed for polyp detection \cite{b12,b13,b14,b15,b16,b17,b18, b19}. In the 2015 MICCAI sub-challenge on automatic polyp detection \cite{b12}, most of the proposed methods were based on CNN, including the winner. The authors in \cite{b13} and \cite{b14} showed that fully convolution network (FCN) architectures could be refined and adapted to recognize polyp structures. Zhang et al. \cite{b15} used FCN-8S to segment polyp region candidates, and texton features computed from each region were used by a random forest classifier for the final decision. Shin et al. \cite{b16} showed that Faster R-CNN is a promising technique for polyp detection. Zhnag et al. \cite{b17} added a tracker to enhance the performance of a CNN polyp detector. Yu et al. \cite{b18} adapted a 3D-CNN model in which a sequence of frames was used for polyp detection. 

In this paper, we adapt Mask R-CNN \cite{b20} for polyp detection and segmentation. Segmenting out polyps from the normal mucosa can help physicians to improve their segmentation errors and subjectivity.  We have several objectives in this study. We first evaluate the performance of Mask R-CNN and compare it to existing methods. Secondly, we aim to evaluate different CNN architectures (e.g., Resnet50 and Resnet101 \cite{b21}, and Inception Resnet V2 \cite{b21}) as the feature extractor for the Mask R-CNN for polyp segmentation. Thirdly, we aim to answer to what extent adding extra training images can help to improve the performance of each of the CNN feature extractors. Do we really need to go for a deeper and more complex CNN to extract higher level of features or do we just need to build a better dataset for training? Finally, we propose an ensemble method for further performance improvement. 

\section{Materials and Methods}

\subsection{Datasets}
Most of the proposed methods mentioned in section \ref{sec:Intro} were tested on different datasets. The authors in \cite{b14,b15} used a dataset containing images of the same polyps for training and testing phases after randomly splitting it into two subsets. This is not very realistic case for validating a method as we may have the same polyps in the training and testing phases. These two issues limit the comparison between the reported results. The 2015 MICCAI sub-challenge on automatic polyp detection was an attempt to evaluate different methods on the same datasets. We, therefore, use the same datasets of 2015 MICCAI polyp detection challenge for training and testing the models. We only use the two datasets of still images: 1) CVC-ClinicDB \cite{b23} containing 32 different polyps  presented in 612 images, and 2) ETIS-Larib \cite{b24} containing 36 different polyps presented in 196 images. In addition, we use CVC-ColonDB \cite{b25} that contains 15 different polyps presented in 300 images.  

\subsection{Evaluation Metrics}
For polyp detection performance evaluation, we calculate recall and precision using the well-known medical parameters such as True Positive (TP), False Positive (FP), True Negative (TN) and False Negative (FN) as follows:
\begin{align}
    recall = \frac{TP}{TP+FN},\\
    precision = \frac{TP}{TP+FP}.
\end{align}
For evaluation of polyp segmentation, we use common segmentation evaluation metrics: Jaccard index (also known as intersection over union, IoU), and  Dice similarity score as follows:
\begin{align}
    J(A,B) = \frac{\mid A \cap B \mid}{\mid A \cup B\mid} = \frac{\mid A \cap B \mid}{\mid A \mid + \mid B \mid - \mid A \cap B \mid},\\
     Dice(A,B) = \frac{2\mid A \cap B \mid}{\mid A \mid + \mid B \mid}, \qquad \qquad \qquad \qquad \quad
\end{align}
where A represents the output image of the method and B the actual ground-truth. 

\subsection{Mask R-CNN}
Mask R-CNN \cite{b20} is a general framework for object instance segmentation. It is an intuitive extension of Faster R-CNN \cite{b26}, the state-of-the-art object detector. Mask R-CNN adapts the same first stage of Faster R-CNN which is region proposal network (RPN). It adds a new branch to the second stage for predicting an object mask in parallel with the existing branches for bounding box regression and confidence value. Instead of using RoIPool, which performs coarse quantization for feature extraction in Faster R-CNN, Mask R-CNN uses RoIAlign, quantization-free layer, to fix the misalignment problem. 
\begin{figure*}[t]
\centering{}\includegraphics[width=18cm,height=7cm]{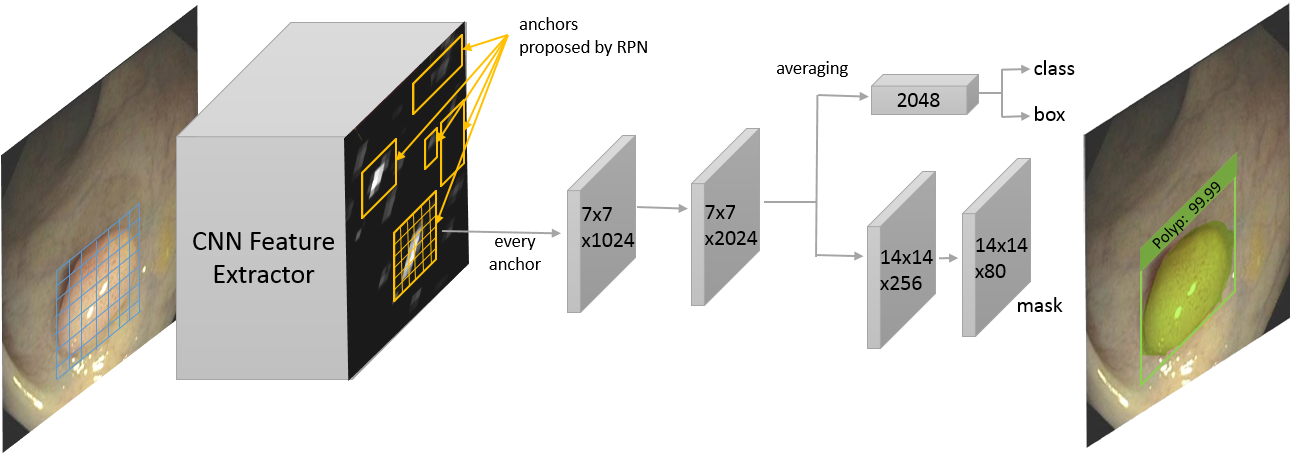}\caption{Our Mask R-CNN framework. In the first stage, we use Resnet50, Resnet101 and Resnet Inception v2 as the feature extractor for the performance evaluation of polyp detection and segmentation. Region proposal network (RPN) utilizes feature maps at one of the intermediate layers (usually the last convolutional layer) of the CNN feature extractor networks to generate box proposals (300 boxes in our study). The proposed boxes are a grid of anchors tiled in different aspect ratios and scales. The second stage predicts the confidence value, the offsets for the proposed box and the mask within the box for each anchor.\label{fig:Mask R-CNN}}
\end{figure*}

For our polyp detection and segmentation, we use the architecture shown in Fig. \ref{fig:Mask R-CNN} to evaluate the performance of Mask R-CNN with different CNN based feature extractors. To train our models, we use a multi-task loss on each region of interest called \textit{anchor} proposed by RPN. For each anchor $a$, we find the best matching ground-truth box $b$. If there is a match, anchor $a$ acts as a positive anchor, and we assign a class label $y_{a}=1$, and a vector ($\phi(b_{a};a)$) encoding box $b$ with respect to anchor $a$. If there is no match, anchor $a$ acts as a negative sample, and the class label is set to $y_{a}=0$. The mask branch has a $14 \times 14$ dimensional output for each anchor. The loss for each anchor $a$, then consists of three losses: location-based loss $\ell_{loc}$ for the predicted box $f_{loc}(I;a,\theta)$, classification loss $\ell_{cls}$ for the predicted class $f_{cls}(I;a,\theta)$ and mask loss $\ell_{mask}$ for the predicted mask $f_{mask}(I,a,\theta)$, where $I$ is the image and $\theta$ is the model parameter, 
\begin{equation}
\hspace*{-0.4225cm} 
    \begin{aligned}
        \mathcal{L}(a,I;\theta)=\frac{1}{m}\sum_{i=1}^{m}\frac{1}{N}\sum_{j=1}^{N}\mathds{1}[a\,is\,positive]\,.\,\ell_{loc}\Big(\phi(b_{a};a)\\
        -f_{loc}(I;a,\theta)\Big)+\ell_{cls}\Big(y_{a},f_{cls}(I;a,\theta)\Big)\qquad \\
        +\ell_{mask}\Big(mask_{a},f_{mask}(I,a,\theta)\Big),\qquad\qquad\:\:
    \end{aligned}
\label{eq:loss function}
\end{equation}where $m$ is the size of mini-batch and $N$ is the number of anchors for each frame. We use the following loss functions: Smooth L1 for the localization loss, softmax for the classification loss and binary cross-entropy for the mask loss.

\subsection{CNN Feature Extractor Networks}
In the first stage of Mask R-CNN, we need a CNN based feature extractor to extract high level features from the input image. The choice of the feature extractor is essential because the CNN architecture, the number of parameters and type of layers directly affect the speed, memory usage and most importantly the performance of the Mask R-CNN. In this study, we select three feature extractors to compare and evaluate their performance in polyp detection and segmentation. We select a deep CNN (e.g., Resnet50 \cite{b21}), deeper CNN (e.g., Resnet101 \cite{b21}), and complex CNN (e.g.,  Inception Resnet (v2) \cite{b22}). 

Resnet is a residual learning framework to ease the training of substantially deep networks to avoid degradation problem--accuracy gets saturated and then degrades rapidly with depth increasing \cite{b21}. With residual learning, we can now benefit from deeper CNN networks to obtain even higher level of features which are essential for difficult tasks such as polyp detection and segmentation. With inception technique, we can increase the depth and width of a CNN network without increasing the computational cost \cite{b27}. Szegedy et al. \cite{b22} proposed Inception Resnet (v2) to combine the optimization benefits of residual learning and computational efficiency from inception units.   

For all three feature extractors, it is important to choose one of the layer to extract features for predicting region proposals by RPN. In our experiments, we use the recommended layers by the original papers. For both Resnet50 and Resnet101, we use the last layer of the $conv4$ block. For Inception Resnet (v2), we use $Mixed\_6a$ layer and its associated residual layers. 

\subsection{Ensemble Model}
The three CNN feature extractors compute different types of features due to differences in their number of layers and architectures. A deeper CNN can compute a higher level of features from the input image while it loses some spatial information due to the contraction and pooling layers. Some polyps might be missed by one of the CNN model while it could be detected by another one. To partly solve this problem, we propose an ensemble model to combine results of two Mask R-CNN models with two different CNN feature extractors. We use one of the models as the main model and its output is always relied on, and the second model as an auxiliary model to support the main model. We only take into account the outputs from the auxiliary model when the confidence of the detection is $\geqslant95\%$ (an optimized value using a validation dataset, see section \ref{Ensamble}). 

\subsection{Training Details}
The available polyp datasets are not large enough to train a deep CNN. To prevent the models from overfitting, we enlarge the dataset by applying different augmentation strategies. We follow the same augmentation methods recommended by Shin et al. \cite{b16}. Image augmentation cannot improve data distribution of the training set---they can only lead to an image-level transformation through depth and scale. This does not ensure the model from being overfitted. Therefore, we use transfer learning by initializing the weights of our CNN feature extractors from models pre-trained on Microsoft\textquoteright s COCO dataset \cite{b28}. We use SGD with a momentum of 0.9, learning rate of 0.0003, and batch size of 1 to fine-tune the pre-trained CNNs using the augmented dataset. We keep the original image size during both training and test phases. 

\section{Results and Discussion} \label{results}

\subsection{Performance Evaluation of the CNN Feature Extractors}
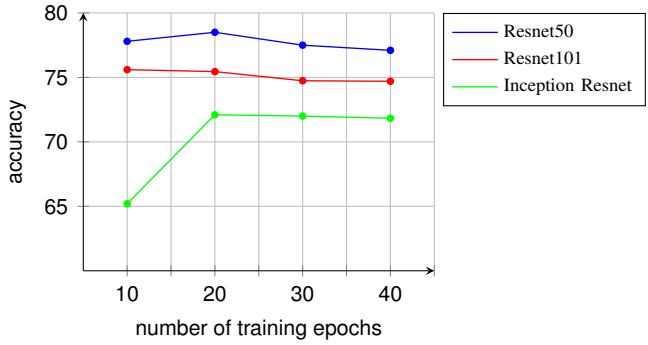
\begin{figure}[ht]
\begin{centering}
\begin{tikzpicture}[font=\footnotesize\sansmath\sffamily] 
\tikzstyle {smallDot} = [ fill=blue, circle, scale=0.3 ] 
\tikzstyle {smallDotR} = [ fill=red, circle, scale=0.3 ] 
\tikzstyle {smallDotG} = [ fill=green, circle, scale=0.3 ] 
\begin{axis} 
[domain = 0:100,
axis on top=false,
axis x line=middle,
axis y line=middle,
xlabel = number of training epochs,
xlabel near ticks,
ylabel = accuracy, 
ylabel near ticks,
width=6.2cm,
height=5cm,
clip  = true,
xmin = 5,  
xmax = 45,
ymin = 60,
ymax = 80,
legend style={font=\scriptsize, at={(1.025,1)},anchor=north west,legend columns=1}, 
grid=major,
xtick={5,10,15,20,25,30,35,40,45},
xticklabels={$\:$,$10$,$\:$,$20$,$\:$,$30$,$\:$,$40$,$\:$,},
ytick={60,65,70,75,80},
yticklabels={$60$,$65$,$70$,$75$,$80$},]  
\addplot [blue, line width = 0.5]
coordinates { (10,77.8) (20,78.5) (30,77.5) (40,77.1)}; 
\addlegendentry[right]{Resnet50};
\node [smallDot] at (axis cs: 10,77.8) {};
\node [smallDot] at (axis cs: 20,78.5) {};
\node [smallDot] at (axis cs: 30,77.5) {};
\node [smallDot] at (axis cs: 40,77.1) {};
\addlegendentry[right]{Resnet101};
\addplot [red, line width = 0.5]
coordinates { (10,75.6) (20,75.45) (30,74.74) (40,74.70)  };     
\node [smallDotR] at (axis cs: 10,75.6) {};
\node [smallDotR] at (axis cs: 20,75.45) {};
\node [smallDotR] at (axis cs: 30,74.74) {};
\node [smallDotR] at (axis cs: 40,74.70) {};
\addplot [green, line width = 0.5]
coordinates { (10,65.2) (20,72.1) (30,72) (40,71.83)};     
\node [smallDotG] at (axis cs: 10,65.2) {};
\node [smallDotG] at (axis cs: 20,72.1) {};
\node [smallDotG] at (axis cs: 30,72) {};
\node [smallDotG] at (axis cs: 40,71.83) {};
\addlegendentry[right]{Inception Resnet}
\end{axis} 
\end{tikzpicture}
\par\end{centering}
\vspace{-3mm}
\caption{Accuracy of the CNN feature extractors vs. number of epochs}
\label{fig2}
\end{figure}

In this section, we report the performance of our Mask R-CNN model shown in Fig. \ref{fig:Mask R-CNN} with the three CNN feature extractors as the base networks. In this experiment, we used CVC-ColonDB for training and CVC-ClinicDB for testing. We trained the three Mask R-CNN models for 10, 20, and 30 epochs and drew curves to show the performance improvement (see Fig. \ref{fig2}). 
We noticed that only 20 epochs was enough to fine-tune the parameters of the three Mask R-CNN models for polyp detection and segmentation, in case of Resnet50 and Resnet101 only 10 epochs. It seems that the the models are getting overfitted on the training dataset after 30 epochs, which results in performance degradation. 

For comparison, we chose 20 epochs and summarized the results in Table \ref{table1}. Inception Resnet (v2) and Resnet101 have shown the best performance for many object classification, detection and segmentation tasks on datasets of natural images \cite{b29}. However, Mask R-CNN with Resnet50 could outperform the counterpart models in all evaluation metrics, with a recall of 83.49\%, precision of 92.95\%, dice of 71.6\% and jaccard of 63.9\%. This might be due to the fact that deeper and more complex networks need larger number of images for training. The CVC-ColonDB dataset contains 300 images with only 15 different polyps. This dataset might not have enough unique polyps for Resnet101 and Inception Resnet (v2) to show their actual performance. This outcome is important because it could be used as evidence to properly choose a CNN feature extractor according to the size of the available dataset. 
\vspace{-2mm}
\begin{table}[htbp]
\caption{Comparison of the results obtained on the CVC-ClinicDB after the models have been trained for 20 epochs}
\begin{center}
\vspace{-1mm}
\begin{tabular}{lccccc}
\hline
\textbf{Mask R-CNNs} & \textbf{\textit{Recall \%}} & \textbf{\textit{Precision \%}}&\textbf{\textit{Dice \%}} & \textbf{\textit{Jaccard \%} }\\
\hline 
{Resnet50} & \textbf{83.49} & \textbf{92.95} & \textbf{71.6} & \textbf{63.9} \\
{Resnet101} & {80.71} & {92.1} & {70.42} & {63.3}\\
{Inception Resnet} & {77.31} & {91.25} & {70.31} & {63.6}\\
\hline
\end{tabular}
\label{table1}
\end{center}
\end{table}
\vspace{-2mm}

\begin{figure}[ht]
\begin{centering}
{\footnotesize{}%
\begin{tabular}{l}
\begin{turn}{90}\footnotesize\sansmath\sffamily{\;\;\; Ground Truth}\end{turn}
\includegraphics[width=1.06in,height=0.9in]{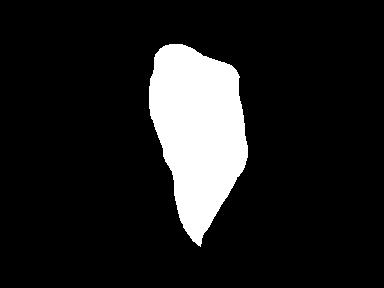}
\includegraphics[width=1.06in,height=0.9in]{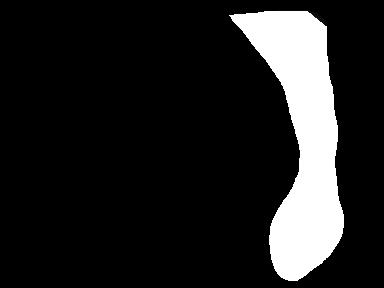}
\includegraphics[width=1.06in,height=0.9in]{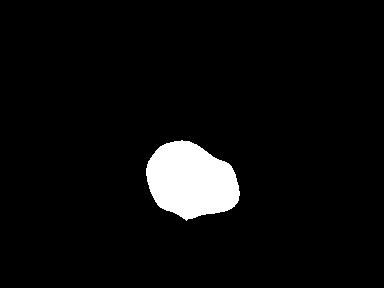}\tabularnewline

\begin{turn}{90}\footnotesize\sansmath\sffamily{\;\;\;\; Input Image}\end{turn}
\includegraphics[width=1.05in,height=0.9in]{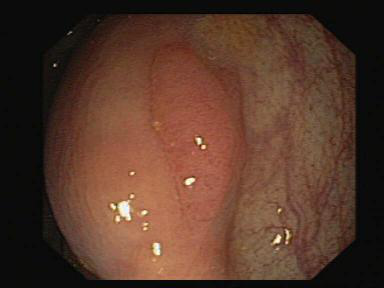}
\includegraphics[width=1.06in,height=0.9in]{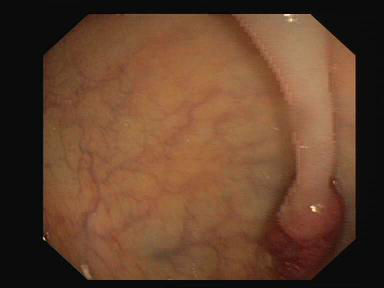}
\includegraphics[width=1.06in,height=0.9in]{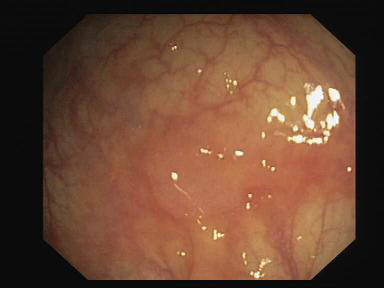}\tabularnewline

\begin{turn}{90}\footnotesize\sansmath\sffamily{\;\;\;\;\; Resnet50}\end{turn}
\includegraphics[width=1.06in,height=0.9in]{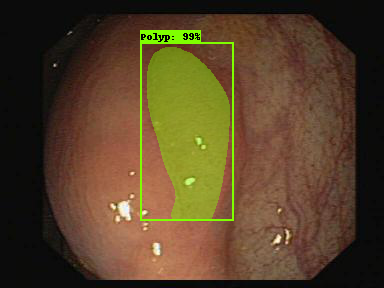}
\includegraphics[width=1.06in,height=0.9in]{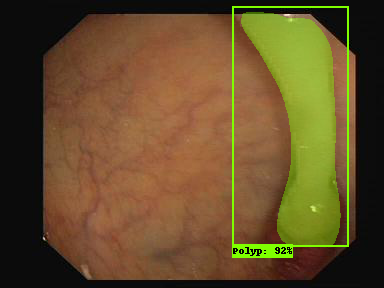}
\includegraphics[width=1.06in,height=0.9in]{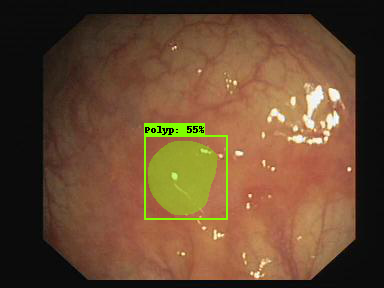}\tabularnewline

\begin{turn}{90}\footnotesize\sansmath\sffamily{\;\;\;\; Resnet101}\end{turn}
\includegraphics[width=1.06in,height=0.9in]{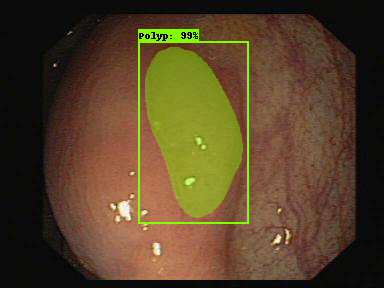}
\includegraphics[width=1.06in,height=0.9in]{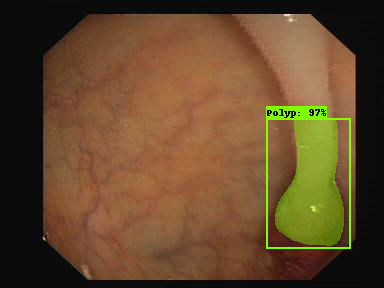}
\includegraphics[width=1.06in,height=0.9in]{Fig1/202_101_Inc.png}\tabularnewline

\begin{turn}{90}\footnotesize\sansmath\sffamily{Inception Resnet}\end{turn}
\includegraphics[width=1.05in,height=0.9in]{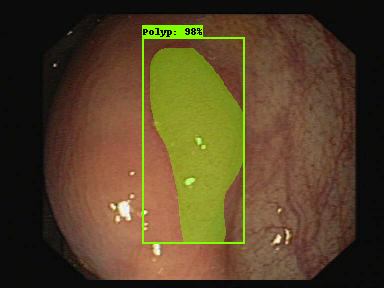}
\includegraphics[width=1.06in,height=0.9in]{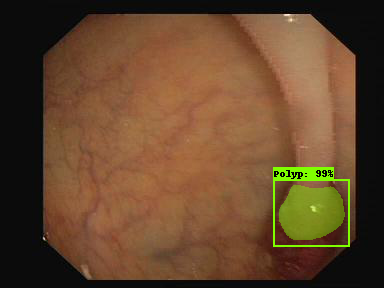}
\includegraphics[width=1.06in,height=0.9in]{Fig1/202_101_Inc.png}\tabularnewline

\end{tabular}
{\footnotesize\par}}
\par\end{centering}
\caption{Example of three outputs produced by our Mask R-CNN models. The images in the 1\textsuperscript{st} row show the ground truths for the polyps shown in the 2\textsuperscript{nd} raw. The images in the 3\textsuperscript{rd} row show the output results produced by Mask R-CNN with Resnet50. The images in the 4\textsuperscript{th} row are outputs from Mask R-CNN with Resnet101. The images in the 5\textsuperscript{th} row are outputs from Mask R-CNN with Resnet Inception (v2).}
\label{fig3_examples}
\end{figure}

Fig. \ref{fig3_examples} illustrates three examples with different output results. The polyp shown in the first column is correctly detected and nicely segmented by the three models. The polyp in the second column is detected correctly by the three models, but only Resnet50 was successful to segment out most of the polyp pixels from the background. The polyp in the third column is only detected and segmented by Resnet50. 

\subsection{Ensemble Results} \label{Ensamble}
It is important to know if detection and segmentation performance can be improved by combining the output results of two Mask R-CNN models. \begin{table}[htbp]
\caption{Ensemble results obtained on the CVC-ClinicDB by combining the results of two Mask R-CNN models}
\begin{center}
\begin{tabular}{lccccc}
\hline
\textbf{Mask R-CNNs} & \textbf{\textit{Recall \%}} & \textbf{\textit{Precision \%}}&\textbf{\textit{Dice \%}} & \textbf{\textit{Jaccard \%} }\\
\hline 
{Resnet50} & {83.49} & {92.95} & {71.6} & {63.9} \\
{Resnet101} &  {80.71} & {92.1} & {70.42} & {63.3} \\
{Resnet Inception} & {77.31} & {91.25} & {70.31} & {63.6} \\
\hline
{Ensemble\textsuperscript{50+101}} & {86.42} & {92.41} & {75.72} & {68.28}\\
{Improvement } & {2.93} & {-0.54} & {4.12} & {4.38}\\
\hline
{Ensemble\textsuperscript{50+Incep}} & {83.95} & {90.67} & {74.73} & {67.41}\\
{Improvement } & {0.46} & {-2.28} & {3.13} & {3.51}\\
\hline
\multicolumn{5}{l}{\textsuperscript{\textbf{50+101}} \;\;Resnet50 used as main, Resnet101 used as auxiliary} \\
\multicolumn{5}{l}{\textsuperscript{\textbf{50+Incep}} Resnet50 used as main, Resnet Inception used as auxiliary} \\
\end{tabular}
\end{center}
\label{table3}
\end{table}\vspace{-3mm}Table \ref{table3} shows the results of this combination. We chose Resnet50 as our main model because it performed better than its counterparts as seen in Table \ref{table1}, and the two others as the auxiliary model. We first used the ETIS-Larib dataset as the validation set to select a suitable confidence threshold  for the auxiliary model. This is an essential prepossessing to prevent increasing the number of FP detection. Based on this optimization step, the output of the auxiliary model is only taken into account when  the confidence of the detection is $\geqslant95\%$.

Table \ref{table3} demonstrates that the auxiliary model could only add a small improvement in the performance of the main model. Resnet101 could improve recall by $2.93\%$, dice by $4.12\%$, and jaccard by $4.38\%$ whereas Resnet Inception could only improve recall by $0.46\%$, dice by $3.13\%$, and jaccard by $3.51\%$. Precision got decreased in both cases. \begin{figure}[ht]
\vspace{-1mm}\begin{centering}
{\footnotesize{}%
\begin{tabular}{l}
{\scriptsize{Input+Ground Truth} \hspace{0.7cm}{Resnet50}  \hspace{1cm}{Resnet101} \hspace{0.6cm} {Recent Inception}}\tabularnewline
\includegraphics[width=0.825in,height=0.7in]{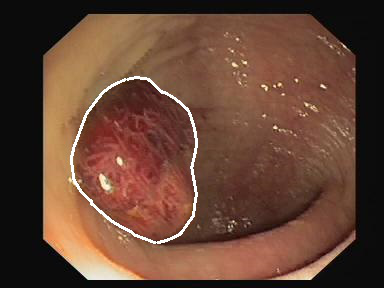}
\includegraphics[width=0.825in,height=0.7in]{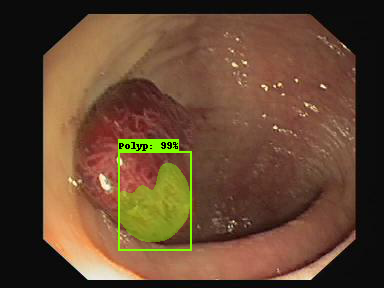}
\includegraphics[width=0.825in,height=0.7in]{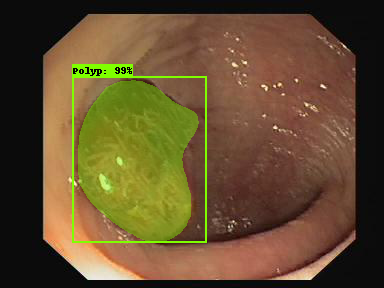}
\includegraphics[width=0.825in,height=0.7in]{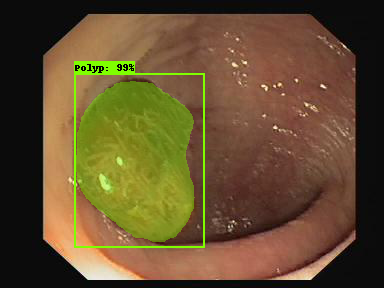}\tabularnewline

\includegraphics[width=0.825in,height=0.7in]{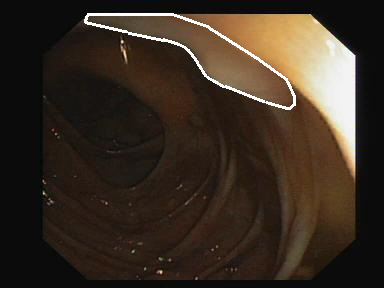}
\includegraphics[width=0.825in,height=0.7in]{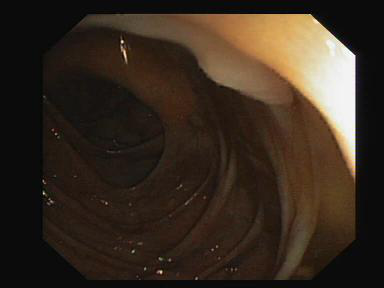}
\includegraphics[width=0.825in,height=0.7in]{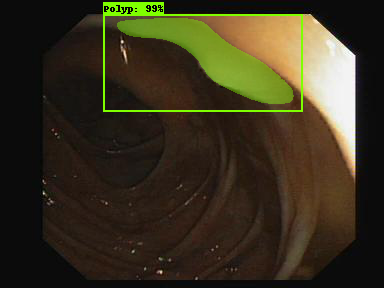}
\includegraphics[width=0.825in,height=0.7in]{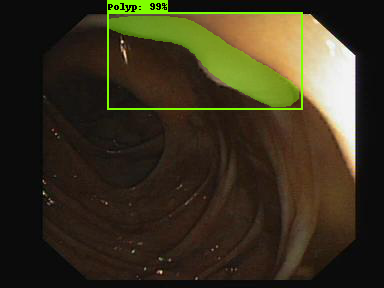}\tabularnewline
\end{tabular}
{\footnotesize\par}}
\par\end{centering}
\caption{Example of two outputs produced by the three Mask R-CNN models. Column 1 shows two polyps with their ground truths. Columns 2, 3 and 4 show the results of Resnet50, Resnet101 and Resnet Inception, respectively.}
\label{fig5_examples}
\end{figure}\vspace{-1mm}The improvement in detection is less than in segmentation. This means that Resnet50 was able to detect most of the polyps detected by the two auxiliary models. Fig. \ref{fig5_examples} illustrates two polyp examples. The first polyp is partially segmented and the second polyp is missed by Resnet50. However, they both are precisely segmented by Resnet101 and Resnet Inception with a confidence of 99\%. 

\subsection{The Effect of Adding New Images to the Training Set}
\begin{figure}[ht]
\begin{centering}
{\footnotesize{}%
\begin{tabular}{l}
\begin{turn}{90}\footnotesize\sansmath\sffamily{\;\;\; Ground Truth}\end{turn}
\includegraphics[width=1.06in,height=0.9in]{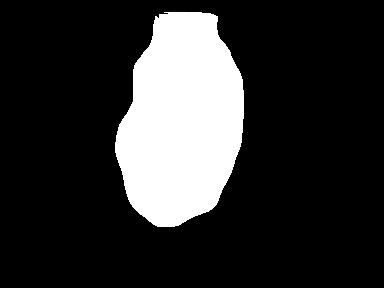}
\includegraphics[width=1.06in,height=0.9in]{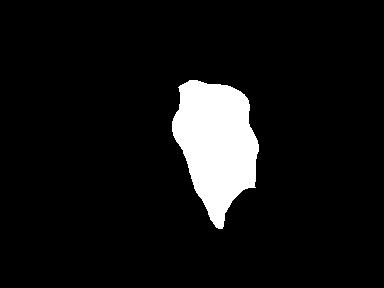}
\includegraphics[width=1.06in,height=0.9in]{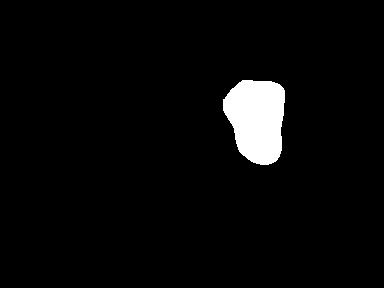}\tabularnewline

\begin{turn}{90}\footnotesize\sansmath\sffamily{\;\;\;\; Input Image}\end{turn}
\includegraphics[width=1.06in,height=0.9in]{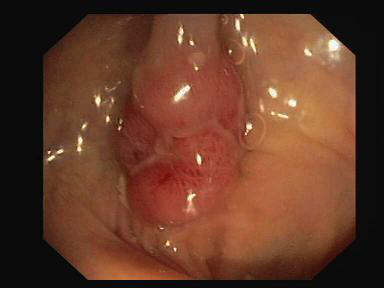}
\includegraphics[width=1.06in,height=0.9in]{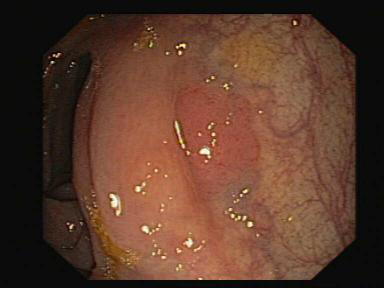}
\includegraphics[width=1.06in,height=0.9in]{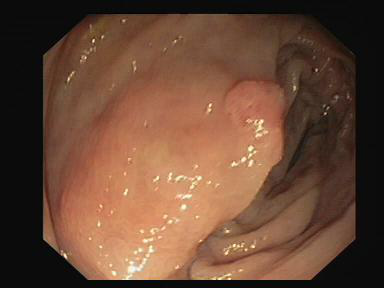}\tabularnewline

\begin{turn}{90}\footnotesize\sansmath\sffamily{\:Resnet Inception\textsuperscript{\textbf*}}\end{turn}
\includegraphics[width=1.06in,height=0.9in]{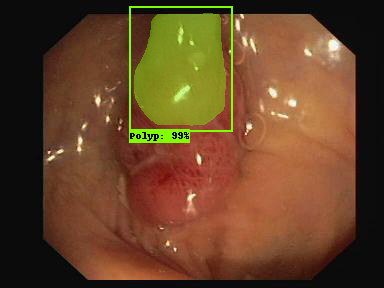}
\includegraphics[width=1.06in,height=0.9in]{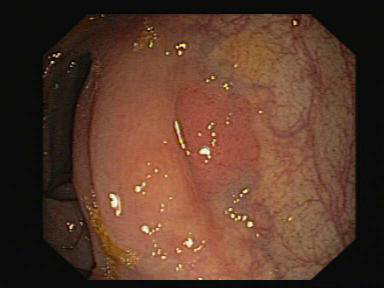}
\includegraphics[width=1.06in,height=0.9in]{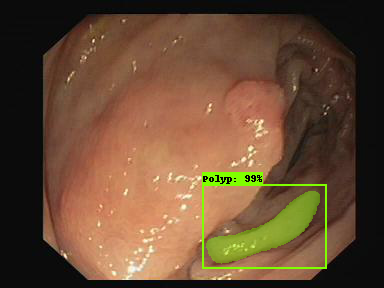}\tabularnewline

\begin{turn}{90}\footnotesize\sansmath\sffamily{Inception Resnet \textsuperscript{\textbf+}}\end{turn}
\includegraphics[width=1.06in,height=0.9in]{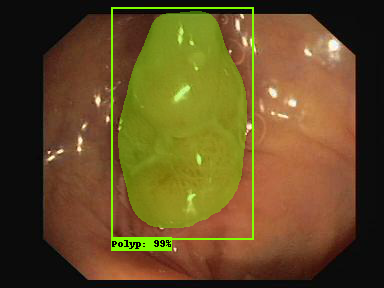}
\includegraphics[width=1.06in,height=0.9in]{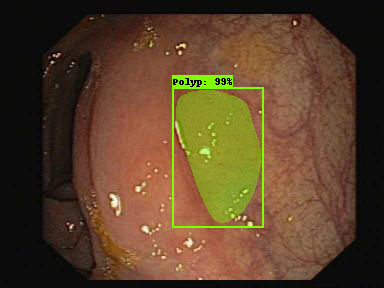}
\includegraphics[width=1.06in,height=0.9in]{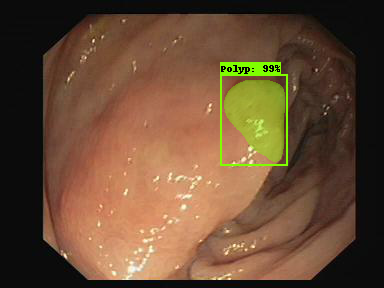}\tabularnewline

\end{tabular}
{\footnotesize\par}}
\par\end{centering}
\caption{Example of three outputs produced by Mask R-CNN with Inception Resnet (v2). The images in the 1\textsuperscript{st} row show the ground truths for the polyps shown in the 2\textsuperscript{nd} row. The images in the 3\textsuperscript{rd} row are output results of the model when trained on CVC-ColonDB (Inception Resnet\textsuperscript*). The images in the 4\textsuperscript{th} row are output results of the model when trained on CVC-ColonDB and ETIS-Larib (Inception Resnet\textsuperscript+).}
\label{fig4_examples}
\end{figure}

In this experiment, we aim to know to what extent adding extra training images with new polyps can help the CNN feature extractors improve their performance. We thus trained the three models again for 20 epochs using the images in both ETIS-Larib and  CVC-ColonDB datasets for training (51 different polyps). Table \ref{table2} shows that all the three models were able to greatly improve both the detection and segmentation capabilities of the Mask R-CNN (especially Inception Resnet) after adding 36 new polyps of ETIS-Larib (196 images) to the training data. Unlike ensemble approach, all the metrics, including precision, improved by larger margins in this experiment. As can be noticed in the results, Resnet Inception is the model with the most improvements in all metrics. This indicates the ability of this CNN architecture to extract richer features from larger training data. 
\begin{table}[htbp]
\caption{Comparison of results obtained on the CVC-ClinicDB after ETIS-Larib was added to the training data and the models trained for 20 epochs}
\vspace{-5mm}
\begin{center}
\begin{tabular}{lccccc}
\hline
\textbf{Mask R-CNNs} & \textbf{\textit{Recall \%}} & \textbf{\textit{Precision \%}}&\textbf{\textit{Dice \%}} & \textbf{\textit{Jaccard \%} }\\
\hline 
{Resnet50\textsuperscript{\textbf*}} & {83.49} & {92.95} & {71.6} & {63.9} \\
{Resnet50\textsuperscript{\textbf+}} & {85.34} & {93.1} & {80.42} & {73.4} \\
{improvement} & {1.85} & {0.15} & {8.82} & {9.5} \\
\hline 
{Resnet101\textsuperscript{\textbf*}} & {80.71} & {92.1} & {70.42} & {63.3}\\
{Resnet101\textsuperscript{\textbf+}} & {84.87} & {95} & {77.48} & {70.13}\\
{improvement} & {4.16} & {2.9} & {7.06} & {6.83} \\
\hline
{Inception Resnet\textsuperscript{\textbf*}} & {77.31} & {91.25} & {70.31} & {63.6}\\
{Inception Resnet\textsuperscript{\textbf+}} & {86.1} & {94.1} & {80.19} & {73.2}\\
{improvement} & {8.79} & {2.85} & {9.88} & {9.6} \\
\hline
\multicolumn{5}{l}{\textsuperscript{\textbf*} indicates that only CVC-ColonDB was used for the training} \\
\multicolumn{5}{l}{\textsuperscript{\textbf+} indicates that CVC-ColonDB and ETIS-Larib were used for training} \\
\end{tabular}
\label{table2}
\end{center}
\end{table}\vspace{-3mm}
As shown in Fig. \ref{fig4_examples}, the new polyp images added to the training data helped Mask R-CNN with Inception Resnet (v2) to predict a better mask for the polyp shown in the first column, correctly detect and segment the missed polyp shown in the second column, and correct the FP detection for the polyp shown in the third column.

\subsection{Comparison with Other Methods}
Each output produced by the Mask R-CNN consists of three components: a confidence value, the coordinates of a bounding box, and a mask (see Fig. \ref{fig3_examples}). This makes Mask R-CNN eligible for performance comparison with other methods in terms of the detection and segmentation capabilities. For comparison against the methods presented in MICCAI 2015, we followed the same dataset guidelines i.e. CVC-ClinicDB dataset used for training stage whereas ETIS-Larib dataset used for testing stage. 
\vspace{-1mm}
\begin{table}[htbp]
\caption{Segmentation Results obtained on the ETIS-Larib dataset}
\begin{center}
\vspace{-1mm}
\begin{tabular}{lccc}\textbf{Segmentation Models} & \textbf{\textit{Dice \%}} & \textbf{\textit{Jaccard \%}} \\
\hline
{FCN-VGG \cite{b13}} & {70.23} & {54.20} \\
\hline 
{Mask R-CNN with Resnet50}  & {58.14} & {51.32} \\
{Mask R-CNN with Resnet101} & \textbf{70.42} & {\textbf{61.24}} \\
{Mask R-CNN with Inception Resnet} & {63.78} & {56.85} \\
\hline
\end{tabular}
\label{table4}
\end{center}
\end{table}
\vspace{-1mm}
In Table \ref{table4}, we compare our Mask R-CNN models against FCN-VGG \cite{b13} which is the only segmentation method fully tested on ETIS-Larib. Our Mask R-CNN with Resnet101 has outperformed all the other methods including FCN-VGG, with a dice of 70.42\% and Jaccard of 61.24\%. To be able to fairly compare the detection capability of our Mask R-CNN models, we followed the same procedure in MICCAI 2015 to compute TP, FP, FN, and TN. As can be seen in Table \ref{table5}, our Mask R-CNN with Resnet101 achieved the highest precision (80\%) and a good recall (72.59\%), outperforming Mask R-CNN with Resnet50, Mask R-CNN with Inception Resnet (v2) and the best method in MICCAI 2015. FCN-VGG has a better recall because both CVC-ClinicDB and ASU-Mayo were used in the training stage (more data for training). These results in Tables \ref{table4} and \ref{table5} are inconsistent with the results in Table \ref{table1} where Resnet50 achieved the best performance. The main reason for this could be due to having more different polyps (32 polyps in 612 images) available for training. 
\begin{table}[htbp]
\caption{Detection Results obtained on the ETIS-Larib dataset}
\begin{center}
\vspace{-1mm}
\begin{tabular}{lccc}
\hline
\textbf{Detection Models} & \textbf{\textit{Recall \%}} & \textbf{\textit{Precision \%}} \\
\hline 
{CUMED \cite{b12}}  & {69.2} & {72.3} \\
{OUS \cite{b12}} & {63.0} & {69.7} \\
{FCN-VGG \cite{b13}} & {\textbf{86.31}} & {73.61} \\
\hline
{Mask R-CNN with Resnet50} & {64.42} & {70.23} \\
{Mask R-CNN with Resnet101} & {72.59} & {\textbf{80.0}} \\
{Mask R-CNN with Inception Resnet} & {64.9} & {77.6} \\
\hline
\end{tabular}
\label{table5}
\end{center}
\end{table}
Again Inception Resnet (v2) was unable to outperform Resnet101. We surmise this is because Inception modules are well-known for being hard to train with a limited amount of training data.   

\section{Conclusions}
In this paper we adapted and evaluated Mask R-CNN with three recent CNN feature extractors i.e. Resnet50, Resnet101, and Inception Resnet (v2) for polyp detection and segmentation. Although a deeper network is essential for high performance in natural image domain, Resnet50 was able to outperform Resnet101 and Resnet Inception (v2) when a limited amount of training data is available. When we added 36 new polyps presented in 196 images to the training data, the three models gained both detection and segmentation improvements, especially for Inception Resnet (v2). The results confirm that with a better training dataset, Mask R-CNN will become a promising technique for polyp detection and segmentation, and using a deeper or more complex CNN feature extractor might become unnecessary.

\end{document}